\documentclass[runningheads]{llncs}
\usepackage{amssymb}
\usepackage{amsmath}
\usepackage{graphicx}
\usepackage{graphics}
\usepackage{adjustbox}
\usepackage{lineno}
\usepackage{color}
\usepackage{algpseudocode}
\usepackage{algorithm}

\algnewcommand\algorithmicinput{\textbf{Input:}}
\algnewcommand\Input{\item[\algorithmicinput]}
\algnewcommand\algorithmicoutput{\textbf{Output:}}
\algnewcommand\Output{\item[\algorithmicoutput]}

\begin{document}
			
\title{An application of a pseudo-parabolic modeling to texture image recognition\thanks{Supported by S\~ao Paulo Research Foundation (FAPESP), National Council for Scientific and Technological Development, Brazil (CNPq), and PETROBRAS - Brazil.}}

\author{Joao B. Florindo\inst{1}\orcidID{0000-0002-0071-0227} \and
	Eduardo Abreu\inst{1}\orcidID{0000-0003-1979-3082}}

\authorrunning{Florindo and Abreu}

\institute{Institute of Mathematics, Statistics and Scientific Computing - University of Campinas,
	Rua S\'{e}rgio Buarque de Holanda, 651, Cidade Universit\'{a}ria "Zeferino Vaz" - Distr. Bar\~{a}o Geraldo, CEP 13083-859, Campinas, SP, Brasil
	\url{http://www.ime.unicamp.br}\\
	\email{\{florindo,eabreu\}@unicamp.br}}

\maketitle

\begin{abstract}
In this work, we present a novel methodology for texture image recognition using a partial differential equation modeling. More specifically, we employ the pseudo-parabolic Buckley-Leverett equation to provide a dynamics to the digital image representation and collect local descriptors from those images evolving in time. For the local descriptors we employ the magnitude and signal binary patterns and a simple histogram of these features was capable of achieving promising results in a classification task. We compare the accuracy over well established benchmark texture databases and the results demonstrate competitiveness, even with the most modern deep learning approaches. The achieved results open space for future investigation on this type of modeling for image ana\-lysis, especially when there is no large amount of data for trai\-ning deep learning models and therefore model-based approaches arise as suitable alternatives.
\end{abstract}

\section{Introduction}

Texture images (also known as visual textures) can be informally defined as those images in which the most relevant information is not encapsulated within one or a limited set of well-defined objects, but rather all pixels share the same importance in their description. This type of image has found numerous application in material sciences \cite{LP19}, medicine \cite{DSBAVJ20}, facial recognition \cite{JZH20}, remote sensing \cite{ZWC20}, cybersecurity \cite{VMS20}, and agriculture \cite{SC20} to name but a few fields with increasing research activity. 

While deep learning approaches have achieved remarkable success in pro\-blems of object recognition and variations of convolutional neural networks have prevailed in the state-of-the-art for this task, texture recognition on the other hand still remains a challenging problem and the classical paradigm of local image encoders still is competitive with the most modern deep neural networks, presenting some advantages over the last ones, like the fact that they can work well even when there is little data available for training. 

In this context, here we present a local texture descriptor based on the action of an operator derived from the Buckley-Leverett partial differential equation (PDE) (see \cite{abreu2017computing,EAPFJV20} and references cited therein). PDE models have been employed in computer vision at least since the 1980's, especially in image processing. The scale-space theory developed by Witkin \cite{W83} and Koenderink \cite{K84} are remarkable examples of such applications.  The anisotropic diffusion equation of Perona and Malik \cite{PM90} also represented great advancement in that research front, as it solved the problem of edge smoothing, common in classical diffusion models. Evolutions of this model were later presented and a survey on this topic was developed in \cite{W97}.

Despite these applications of PDEs in image processing, substantially less research has been devoted to recognition. As illustrated in \cite{vieira2020texture}, pseudo-parabolic PDEs are promising models for this purpose. An important characteristic of these models is that jump discontinuities in the initial condition are replicated in the solution \cite{cuesta2009numerical}. This is an important feature in recognition as it allows some control over the smoothing effect and would preserve relevant edges, which are known to be very important in image description.

Based on this context, we propose the use of Buckley-Leverett equation as an operator acting as a nonlinear filter over the texture image. That image is used as initial condition for the PDE problem and the solution obtained by a numerical scheme developed in \cite{abreu2017computing} is used to compose the image representation. The solution at each time is encoded by a local descriptor. Extending the idea presented in \cite{vieira2020texture}, here we propose two local features: the sign and the magnitude binary patterns \cite{GZZ10b}. The final texture descriptors are provided by simple concatenation of histograms over each time.

The effectiveness of the proposed descriptors is validated on the classification of well established benchmark texture datasets, more exactly, KTH-TIPS-2b \cite{HCFE04} and UIUC \cite{LSP05}. The accuracy is compared with the state-of-the-art in texture recognition, including deep learning solutions, and the results demonstrate the potential of our approach, being competitive with the most advanced solutions recently published on this topic.    

\section{Partial differential equation and numerical modeling}\label{sec:pde}

We consider an advanced simulation approach for the 
pseudo-parabolic PDE
\begin{equation}
	\frac{\partial u}{\partial t} 
	= \nabla \cdot \mathbf{w},
	\qquad 
	\text{ where }
	\qquad 
	\mathbf{w} =   
	g(x,y,t) \, \nabla \left(u 
	+ \tau \,  \frac{\partial u}{\partial t} \right),
	\label{eq:pptransp2}
\end{equation}
and let $\Omega \subset \mathbb{R}^{2}$ denote a rectangular 
domain and 
${\displaystyle u( \,\cdot\,,\, \cdot \, , \,t):\Omega 
\rightarrow \mathbb {R} }$
be a sequence of images that satisfies the pseudo-parabolic 
equation \eqref{eq:pptransp2}, in which the original image at 
$t = 0$ corresponds to the initial condition, along with zero 
flux condition across the domain boundary $\partial \Omega$,
$\mathbf{w} \cdot \mathbf{n} = 0, \, (x \,,\, y) \in \partial \Omega$.
Following \cite{vieira2020texture} (see also \cite{VA18,abreu2017computing}),
we consider the discretization modeling of the PDE (\ref{eq:pptransp2}) 
in a uniform partition of $\Omega$ into rectangular subdomains
$\Omega_{i,j}$, for $i = 1 \,,\, \dots \,,\, m$ and $j = 1 \,,\, \dots \,,\, l$, 
with dimensions $\Delta x \times \Delta y$. The center of each 
subdomain $\Omega_{i,j}$ is denoted by $(x_i \,,\, y_j)$. Given 
a final time of simulation $T$, consider a uniform partition of 
the interval $[0 \,,\, T]$ into $N$ subintervals, where the time 
step $\Delta t = T/N$ is subject to a stability condition (see 
\cite{vieira2020texture,VA18} for details). We denote the texture 
configuration frames in the time levels $t_n = n \Delta t$, for 
$n = 0 \,,\, \dots \,,\, N$. Let $U_{i,j}^n$ and $W_{i,j}^{n+1}$
be a finite difference approximations for $u(x_i \,,\, y_j \,,\, t_n)$
and $\mathbf{w}$, respectively, and both related to the pseudo-parabolic 
PDE modeling of \eqref{eq:pptransp2}. Motivated by our promising 
results in \cite{vieira2020texture}, we employ a stable cell-centered 
finite difference discretization in space after applying the backward 
Euler method in time to \eqref{eq:pptransp2}, yielding 
\begin{equation}
	\frac{U_{i,j}^{n+1} - U_{i,j}^n}{\Delta t}
	=
	\frac{W_{i+\frac{1}{2},j}^{n+1} - W_{i-\frac{1}{2},j}^{n+1}}{\Delta x}
	+ \frac{W_{i,j+\frac{1}{2}}^{n+1} - W_{i,j-\frac{1}{2}}^{n+1}}{\Delta y}.
\label{disPB}
\end{equation}
Depending on the application as well as the calibration data 
and texture para\-meters upon model \eqref{eq:pptransp2}, we will 
have linear or nonlinear diffusion models for image processing 
(see, e.g., \cite{PM90,CLMC92,W97}). As a result of this process 
the discrete problem (\ref{disPB}) would be linear-like 
$\mathbf{A}^n \, \mathbf{U}^{n+1} = \mathbf{b}^n$ or nonlinear-like 
$\mathbf{F}(\mathbf{U}^{n+1}) = 0$ and several interesting methods 
can be used (see, e.g., \cite{EAPFJV20,vieira2020texture,VA18,PM90,CLMC92,W97}). 
We would like to point out at this moment that our contribution relies 
on the PDE modeling of \eqref{eq:pptransp2} as well as on the 
calibration data and texture parameters associated to the 
pseudo-parabolic modeling in conjunction with a fine tunning of 
the local descriptors for texture image recognition for the 
pertinent application under consideration. In summary, we have 
a family of parameter choice strategies that combines 
pseudo-parabolic modeling with texture image recognition.

Here, we consider the diffusive flux as 
$g(x \,,\ y \,,\, t) \equiv 1$, which results \eqref{eq:pptransp2} 
to be a linear pseudo-parabolic model. For a texture image classification 
based on a pseudo-parabolic diffusion model to be processed, we 
just consider that each subdomain $\Omega_{i,j}$ corresponds to 
a pixel with $\Delta x = \Delta y = 1$. As we perform an implicit 
robust discretization in time (backward Euler), we simply choose 
the time step $\Delta t = \Delta x$ and the damping coefficient 
$\tau = 5$. More details can be found in \cite{vieira2020texture}; 
see also \cite{VA18,abreu2017computing}.

Therefore, this description summarizes the basic key ideas of 
our computational PDE modeling approach for texture image 
classification based on a pseudo-parabolic diffusion model  
(\ref{eq:pptransp2}).

\section{Proposed methodology}

Inspired by ideas presented in \cite{vieira2020texture} and significantly extending comprehension on that study, here we propose the development of a family of images $\{u_k\}_{k=1}^K$. These images are obtained by introducing the original image $u_0$ as initial condition for the 2D pseudo-parabolic numerical scheme presented in Section \ref{sec:pde}. $u_k$ is the numerical solution at each time $t=t_k$. Here, $K=50$ showed to be a reasonable balance between computational performance and description quality.

Following that, we collected two types of local binary descriptors \cite{GZZ10b} from each $u_k$. More exactly, we used sign $LBP\_S_{P,R}^{riu2}$ and magnitude $LBP\_M_{P,R}^{riu2}$ descriptors. In short, the local binary sign pattern $LBP\_S_{P,R}^{riu2}$ for each image pixel with gray level $g_c$ and whose neighbor pixels at distance $R$ have intensities $g_p$ ($p=1,\cdots,P$) is given by
\begin{equation} LBP\_S_{P,R}^{riu2} = \left\{
\begin{array}{ll}
\sum_{p=0}^{P-1}H(g_p-g_c)2^p & \mbox{if } \mathcal{U}(LBP_{P,R})\geq 2\\
P+1 & \mbox{otherwise}, 
\end{array}
\right. \end{equation} 
where $H$ corresponds to the Heaviside step function ($H(x)=1$ if $x\geq 0$ and $H(x)=0$ if $x<0$) and $\mathcal{U}$ is the uniformity function, defined by
\begin{equation} \mathcal{U}(LBP_{P,R}) = |H(g_{P-1}-g_c)-H(g_0-g_c)| + \sum_{p=1}^{P-1}|H(g_{p}-g_c)-H(g_{p-1}-g_c)|. \end{equation}

Similarly, the magnitude local descriptor is defined by

\begin{equation} LBP\_M_{P,R}^{riu2} = \left\{
\begin{array}{ll}
\sum_{p=0}^{P-1}t(|g_p-g_c|,C)2^p & \mbox{ \, if \, } \mathcal{U}(LBP_{P,R})\geq 2\\
P+1 & \mbox{otherwise}, 
\end{array}\right.
\end{equation}
where $C$ is the mean value of $|g_p-g_c|$ over the whole image and $t$ is a threshold function, such that $t(x,c) = 1$ if $x\geq c$ and $t(x,c)=0$, otherwise.

\begin{figure}[!htpb]
	\centering
	\includegraphics[width=\textwidth]{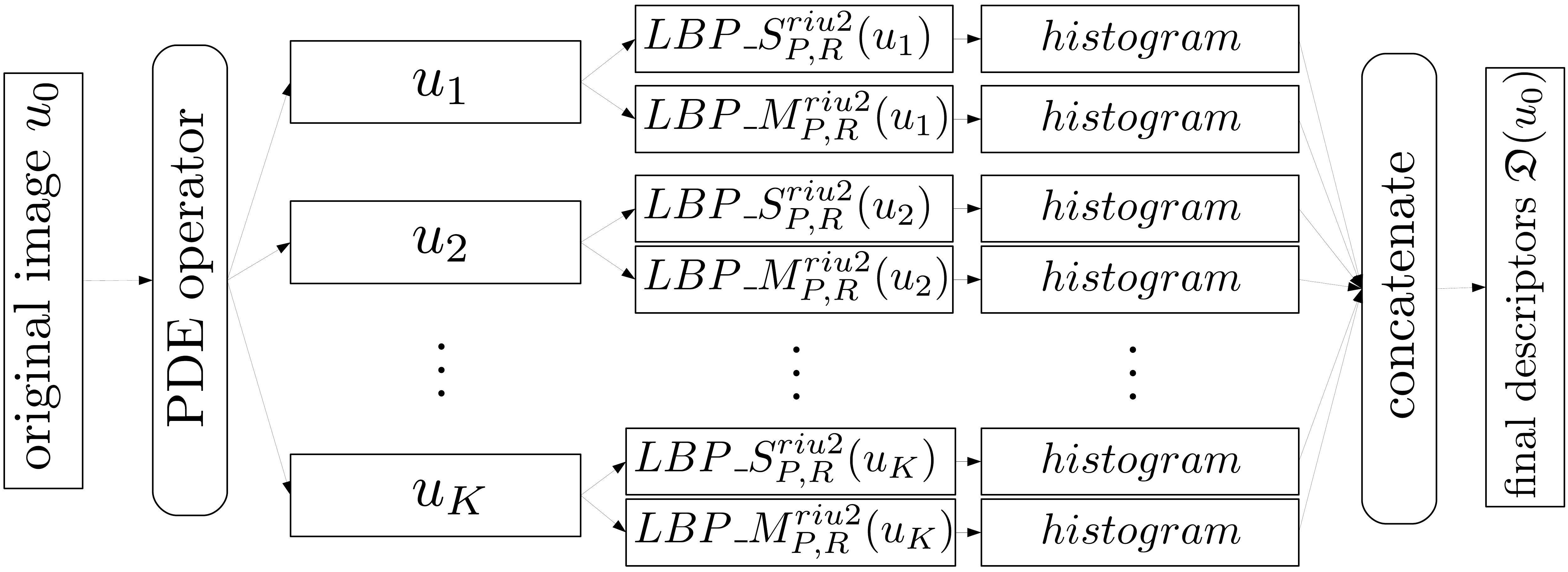}
	\caption{Main steps of the proposed method.}
	\label{fig:method}
\end{figure}

Finally, we compute the histogram $\mathrm{h}$ of $LBP\_S_{P,R}^{riu2}(u_k)$ and $LBP\_M_{P,R}^{riu2}(u_k)$ for the following pairs of $(P,R)$ values: $\{ (1,8),(2,16),(3,24),(4,24) \}$. The proposed descriptors can be summarized by
\begin{equation}
\mathfrak{D}(u_0) = \bigcup_{\mbox{type}=\{S,M\}}\bigcup_{\substack{(P,R) = \{(8,1),\\(16,2),(24,3),\\(24,4)\}}}\bigcup_{k=0}^K \mathrm{h}(LBP\_\mbox{type}_{P,R}^{riu2}(u_k)).
\end{equation}
To reduce the dimensionality of the final descriptors, we also apply Karhunen-Lo\`{e}ve transform \cite{P1901} before their use as input to the classifier algorithm. The diagram depicted in Figure \ref{fig:method} illustrates the main steps involved in the proposed algorithm.

\section{Experiments}

The performance of the proposed descriptors is assessed on the classification of two well-established benchmark datasets of texture images, namely, KTH-TIPS-2b \cite{HCFE04} and UIUC \cite{LSP05}.

KTH-TIPS-2b is a challenging database focused on the real material depicted in each image rather than on the texture instance as most classical databases. In this way, images collected under different configurations (illumination, scale and pose) should be part of the same class. The database comprises a total of 4752 color textures with resolution $200 \times 200$ (here they are converted to gray scales), equally divided into 11 classes. Each class is further divided into 4 samples (each sample corresponds to a specific configuration). We adopt the most usual (and most challenging) protocol of using one sample for training and the remaining three samples for testing.

UIUC is a gray-scale texture dataset composed by 1000 images with resolution $256\times 256$ evenly divided into 25 classes. The images are photographed under uncontrolled natural conditions and contain variation in illumination, scale, pers\-pective and albedo. For the training/testing split we also follow the most usual protocol, which consists in half of the images (20 per class) randomly selected for training and the remaining half for testing. This procedure is repeated 10 times to allow the computation of an average accuracy.

For the final step of the process, which is the machine learning classifier, we use Linear Discriminant Analysis \cite{F36}, given its easy interpretation, absence of hyper-parameters to be tuned and known success in this type of application \cite{vieira2020texture}. 

\section{Results and Discussion}

\begin{figure}[!htpb]
	\centering
	\includegraphics[width=.8\textwidth]{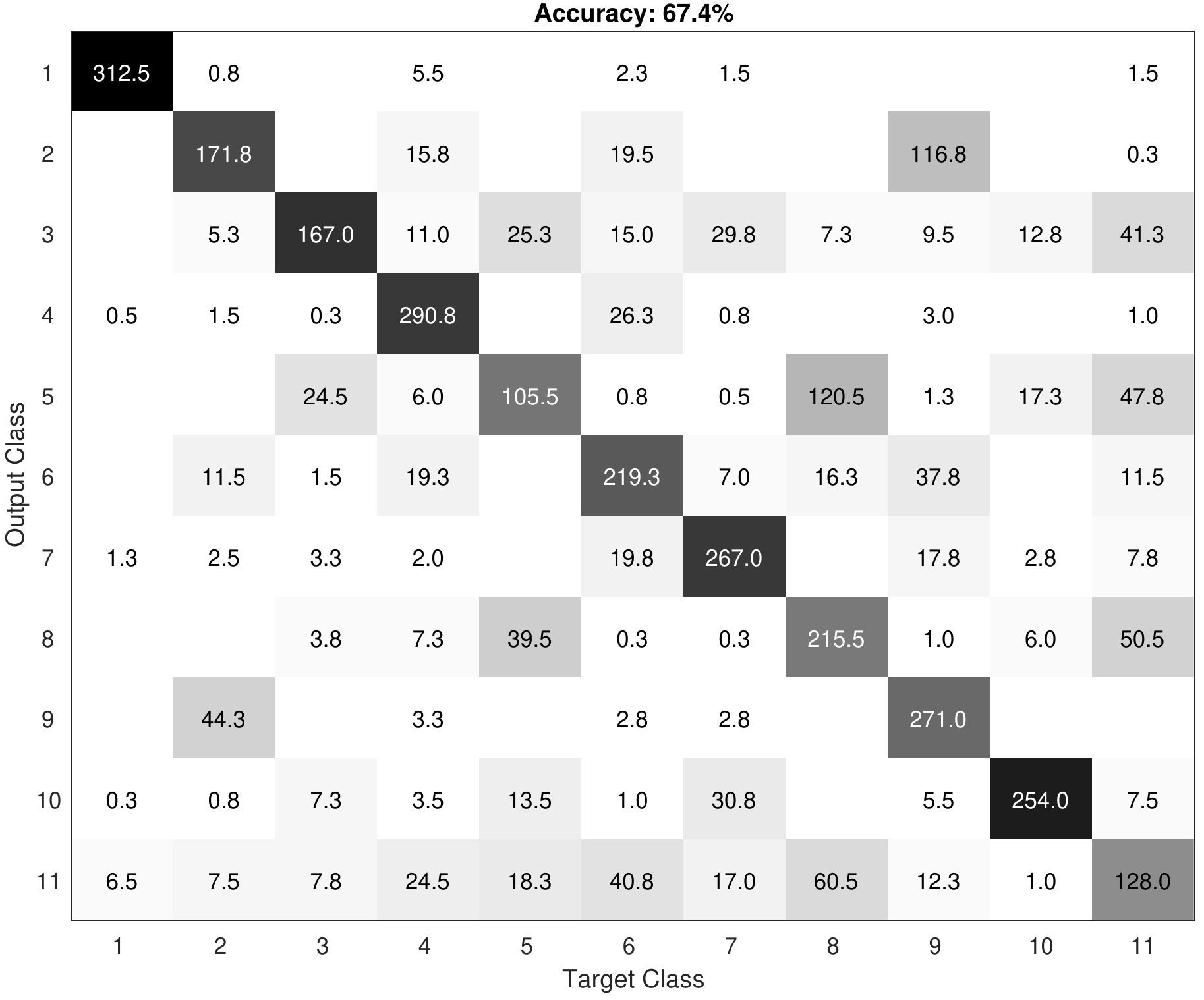}
	\caption{Average confusion matrix and accuracy for KTH-TIPS-2b.}
	\label{fig:CM_kthtips2b}
\end{figure}

\begin{figure}[!htpb]
	\centering
	\includegraphics[width=.8\textwidth]{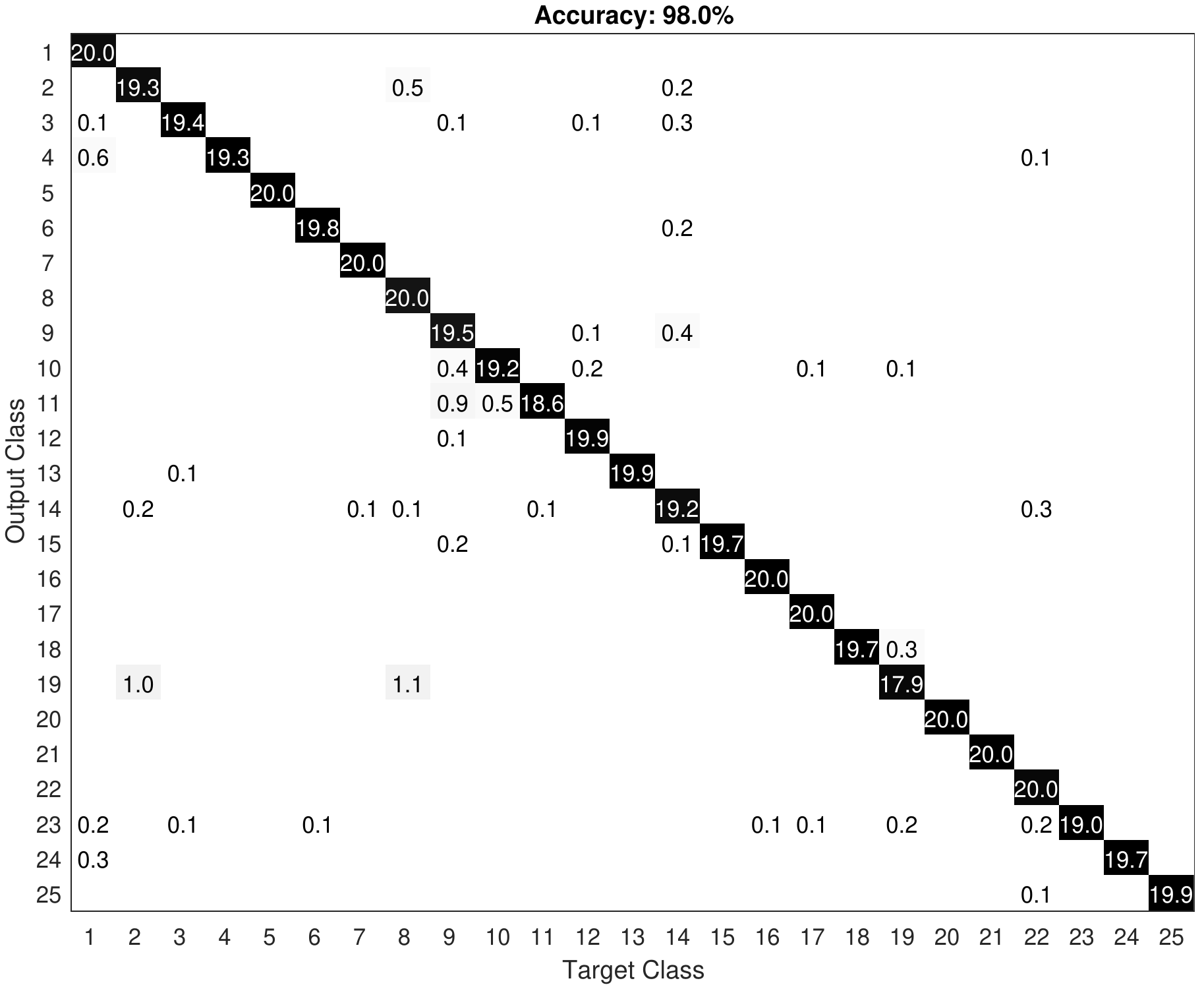}
	\caption{Average confusion matrix and accuracy for UIUC.}
	\label{fig:CM_uiuc}
\end{figure}

Figures \ref{fig:CM_kthtips2b} and \ref{fig:CM_uiuc} show the average confusion matrices and accuracies (percentage of images correctly classified) for the proposed descriptors in the classification of KTH-TIPS-2b and UIUC, respectively. The average is computed over all trai\-ning/testing rounds, corresponding, respectively, to 4 rounds in KTH-TIPS-2b and 10 rounds in UIUC. This is an interesting and intuitive graphical representation of the most complicated classes and the most confusable pairs of classes. While in UIUC there is no pair of classes deserving particular attention (the maximum confusion is of one image), KTH-TIPS-2b exhibits a much more challenging scenario. The confusion among classes 3, 5, 8, and 11 is the most cri\-tical scenario for the proposed classification framework. It turns out that these classes correspond, respectively, to the materials ``corduroy'', ``cotton'', ``linen'', and ``wool''. Despite being different materials, they inevitably share similarities as at the end they are all types of fabrics. Furthermore, looking at the sample from these classes, we can also observe that the attribute ``color'', that is not considered here, would be a useful class discriminant in that case. In general, the performance of our proposal in this dataset is quite promising and the confusion matrix and raw accuracy confirm our theoretical expectations.

Table \ref{tab:SRdatabase} presents the accuracy compared with other methods in the literature, including several approaches that can be considered as part of the state-of-the-art in texture recognition. First of all, the advantage over the original CLBP, whose part of the descriptors are used here as local encoder, is remarkable, being more than 10\% in KTH-TIPS-2b. Other advanced encoders based on SIFT are also outperformed in both datasets (by a large margin in the most challen\-ging textures of KTH-TIPS-2b). SIFT descriptors are complex object descriptors and were considered the state-of-the-art in image recognition for several years. Compared with the most recent CNN-based approaches presented in \cite{CMKV16}, the results are also competitive. In UIUC, the proposed approach outperforms CNN methods like DeCAF and FC-CNN VGGM. These correspond to complex architectures with a high number of layers and large requirements of computational resources and whose results are pretty hard to be interpreted. 
\begin{table}[!htpb]
	\centering
	\caption{Accuracy of the proposed descriptors compared with other texture descriptors in the literature. A superscript $^1$ in KTHTIPS-2b means training on three samples  and testing on the remainder (no published results for the setup used here).}
	\label{tab:SRdatabase}
	\begin{tabular}{cc}
		KTH-TIPS-2b & UIUC\\
		\begin{tabular}{cc}
			\hline
			Method & Acc. (\%)\\
			\hline
			VZ-MR8 \cite{VZ05} & 46.3\\
			LBP \cite{OPM02} & 50.5\\
			VZ-Joint \cite{VZ09} & 53.3\\
           	BSIF \cite{KR12} & 54.3\\			
			LBP-FH \cite{AMHP09} & 54.6\\
			CLBP \cite{GZZ10b} & 57.3\\
           	SIFT+LLC \cite{CMKV16} & 57.6\\         			
			ELBP \cite{LZLKF12} & 58.1\\
			SIFT + KCB \cite{CMKMV14} & 58.3\\
			SIFT + BoVW \cite{CMKMV14} & 58.4\\
           	LBP$_{riu2}$/VAR \cite{OPM02} & 58.5\footnotemark[1]\\			
           	PCANet (NNC) \cite{CJGLZM15} & 59.4\footnotemark[1]\\			
           	RandNet (NNC) \cite{CJGLZM15} & 60.7\footnotemark[1]\\			
			SIFT + VLAD \cite{CMKMV14} & 63.1\\  	
           	ScatNet (NNC) \cite{BM13} & 63.7\footnotemark[1]\\
           	FV-CNN AlexNet \cite{CMKV16} & 69.7\\
			\hline
			\textbf{Proposed} & \textbf{67.4}\\
			\hline
		\end{tabular}
		&
		\begin{tabular}{cc}
			\hline
			Method & Acc. (\%)\\
			\hline
			RandNet (NNC) \cite{CJGLZM15} & 56.6\\
			PCANet (NNC) \cite{CJGLZM15} & 57.7\\
			BSIF \cite{KR12} & 73.4\\
			VZ-Joint \cite{VZ09} & 78.4\\
			LBP$_{riu2}$/VAR \cite{OPM02} & 84.4\\
			LBP \cite{OPM02} & 88.4\\
			ScatNet (NNC) \cite{BM13} & 88.6\\
			MRS4 \cite{VZ09} & 90.3\\
			SIFT + KCB \cite{CMKMV14} & 91.4\\
			MFS \cite{XJF09} & 92.7\\
			VZ-MR8 \cite{VZ05} & 92.8\\
			DeCAF \cite{CMKMV14} & 94.2\\
			FC-CNN VGGM \cite{CMKV16} & 94.5\\
			CLBP \cite{GZZ10b} & 95.7\\
			SIFT+BoVW \cite{CMKMV14} & 96.1\\
			SIFT+LLC \cite{CMKV16} & 96.3\\			
			\hline
			\textbf{Proposed} & \textbf{98.0}\\
			\hline
		\end{tabular}
	\end{tabular}
\end{table}

Generally speaking, the proposed method provided results in texture classification that confirm its potential as a texture image model. Indeed that was theoretically expected from its ability of smoothing spurious noise at the same time that preserves relevant discontinuities on the original image. The combination with a powerful yet simple local encoder like CLBP yielded interesting and promising performance neither requiring large amount of data for training nor advanced computational resources. In general, such great performance combined with the straightforwardness of the model, that allows some interpretation of the texture representation based on local homo/heterogeneous patterns, make the proposed descriptors a candidate for practical applications in texture analysis, especially when we have small to medium datasets and excessively complicated algorithms should be avoided. 

\section{Conclusions}

In this study, we investigated the performance of a nonlinear PDE model (pseudo-parabolic) as an operator for the description of texture images. The operator was applied for a number of iterations (time evolution) and a local encoder was collected from each transformed image. The use of a basic histogram to pooling the local encoders was sufficient to provide competitive results.

The proposed descriptors were evaluated over a practical task of texture classification on benchmark datasets and the accuracy was compared with other approaches from the state-of-the-art. Our method outperformed several other local descriptors that follow similar paradigm and even some learning-based algorithms employing complex versions of convolutional neural networks.

The obtained results confirmed our expectations of a robust texture descriptor, explained by its ability of nonlinearly smoothing out spurious noise and unnecessary details, but preserving relevant information, especially those conveyed by sharp discontinuities. In general, the results and the confirmation of the theoretical formulation suggest the suitability of applying such model in practice, in tasks of texture recognition that require simple models, easy to be interpreted and that do not require much data for training. This is a common situation in areas like medicine and several others. 

\section*{Acknowledgements}

J. B. Florindo gratefully acknowledges the financial support of S\~ao Paulo Research Foundation (FAPESP) (Grant \#2016/16060-0) and from National Council for Scientific and Technological Development, Brazil (CNPq) (Grants \\\#301480/2016-8 and \#423292/2018-8). E. Abreu gratefully acknowledges the financial support of S\~ao Paulo Research Foundation (FAPESP) (Grant \\\#2019/20991-8), from National Council for Scientific and Technological Deve\-lopment - Brazil (CNPq) (Grant \#2 306385/2019-8) and PETROBRAS - Brazil (Grant \#2015/00398-0). E. Abreu and J. B. Florindo also gratefully acknowledge the financial support of Red Iberoamericana de Investigadores en Matem\'aticas Aplicadas a Datos (MathData).


\begin{thebibliography}{10}
	\providecommand{\url}[1]{\texttt{#1}}
	\providecommand{\urlprefix}{URL }
	\providecommand{\doi}[1]{https://doi.org/#1}
	
	\bibitem{abreu2017computing}
	Abreu, E., Vieira, J.: Computing numerical solutions of the pseudo-parabolic
	buckley--leverett equation with dynamic capillary pressure. Math. Comput.
	Simul.  \textbf{137},  29--48 (2017)
	
	\bibitem{EAPFJV20}
	Abreu, E., Ferraz, P., Vieira, J.: Numerical resolution of a pseudo-parabolic
	buckley-leverett model with gravity and dynamic capillary pressure in
	heterogeneous porous media. Journal of Computational Physics  \textbf{411},
	109395 (2020). \doi{https://doi.org/10.1016/j.jcp.2020.109395},
	\url{http://www.sciencedirect.com/science/article/pii/S0021999120301698}
	
	\bibitem{AMHP09}
	Ahonen, T., Matas, J., He, C., Pietik{\"a}inen, M.: Rotation invariant image
	description with local binary pattern histogram fourier features. In:
	Salberg, A.B., Hardeberg, J.Y., Jenssen, R. (eds.) Image Analysis. pp.
	61--70. Springer Berlin Heidelberg, Berlin, Heidelberg (2009)
	
	\bibitem{BM13}
	Bruna, J., Mallat, S.: Invariant scattering convolution networks. IEEE
	Transactions on Pattern Analysis and Machine Intelligence  \textbf{35}(8),
	1872--1886 (2013)
	
	\bibitem{CLMC92}
	Catt\'{e}, F., Lions, P.L., Morel, J.M., Coll, T.: Image selective smoothing
	and edge detection by nonlinear diffusion. SIAM Journal on Numerical Analysis
	\textbf{29}(1),  182--193 (1992)
	
	\bibitem{CJGLZM15}
	{Chan}, T., {Jia}, K., {Gao}, S., {Lu}, J., {Zeng}, Z., {Ma}, Y.: Pcanet: A
	simple deep learning baseline for image classification? IEEE Transactions on
	Image Processing  \textbf{24}(12),  5017--5032 (2015)
	
	\bibitem{CMKMV14}
	Cimpoi, M., Maji, S., Kokkinos, I., Mohamed, S., Vedaldi, A.: Describing
	textures in the wild. In: Proceedings of the 2014 IEEE Conference on Computer
	Vision and Pattern Recognition. pp. 3606--3613. CVPR '14, IEEE Computer
	Society, Washington, DC, USA (2014)
	
	\bibitem{CMKV16}
	Cimpoi, M., Maji, S., Kokkinos, I., Vedaldi, A.: Deep filter banks for texture
	recognition, description, and segmentation. International Journal of Computer
	Vision  \textbf{118}(1),  65--94 (2016)
	
	\bibitem{cuesta2009numerical}
	Cuesta, C., Pop, I.: Numerical schemes for a pseudo-parabolic burgers equation:
	discontinuous data and long-time behaviour. J. Comput. Appl. Math.
	\textbf{224},  269--283 (2009)
	
	\bibitem{DSBAVJ20}
	Dhivyaa, C.R., Sangeetha, K., Balamurugan, M., Amaran, S., Vetriselvi, T.,
	Johnpaul, P.: {Skin lesion classification using decision trees and random
		forest algorithms}. {Journal of Ambient Intelligence and Humanized Computing}
	({2020})
	
	\bibitem{F36}
	Fisher, R.A.: The use of multiple measurements in taxonomic problems. Annals of
	Eugenics  \textbf{7}(2),  179--188 (1936)
	
	\bibitem{GZZ10b}
	Guo, Z., Zhang, L., Zhang, D.: A completed modeling of local binary pattern
	operator for texture classification. Trans. Img. Proc.  \textbf{19}(6),
	1657–1663 (2010)
	
	\bibitem{HCFE04}
	Hayman, E., Caputo, B., Fritz, M., Eklundh, J.O.: On the significance of
	real-world conditions for material classification. In: Pajdla, T., Matas, J.
	(eds.) Computer Vision - ECCV 2004. pp. 253--266. Springer Berlin Heidelberg,
	Berlin, Heidelberg (2004)
	
	\bibitem{JZH20}
	Jain, D.K., Zhang, Z., Huang, K.: {Multi angle optimal pattern-based deep
		learning for automatic facial expression recognition}. {Pattern Recognition
		Letters}  \textbf{{139}},  {157--165} ({2020})
	
	\bibitem{KR12}
	Kannala, J., Rahtu, E.: Bsif: Binarized statistical image features. In: ICPR.
	pp. 1363--1366. IEEE Computer Society (2012)
	
	\bibitem{K84}
	Koenderink, J.J.: The structure of images. Biological Cybernetics
	\textbf{50}(5),  363--370 (1984)
	
	\bibitem{LSP05}
	Lazebnik, S., Schmid, C., Ponce, J.: A sparse texture representation using
	local affine regions. IEEE Transactions on Pattern Analysis and Machine
	Intelligence  \textbf{27}(8),  1265--1278 (2005)
	
	\bibitem{LP19}
	Lin, J., Pappas, T.N.: {Structural texture similarity for material
		recognition}. In: {2019 IEEE International Conference on Image Processing
		(ICIP)}. pp. {4424--4428}. {IEEE International Conference on Image Processing
		ICIP}, {Inst Elect \& Elect Engineers; Inst Elect \& Elect Engineers Signal
		Proc Soc} ({2019}), {26th IEEE International Conference on Image Processing
		(ICIP), Taipei, TAIWAN, SEP 22-25, 2019}
	
	\bibitem{LZLKF12}
	Liu, L., Zhao, L., Long, Y., Kuang, G., Fieguth, P.: Extended local binary
	patterns for texture classification. Image Vision Comput.  \textbf{30}(2),
	86--99 (2012)
	
	\bibitem{OPM02}
	Ojala, T., Pietik\"{a}inen, M., M\"{a}enp\"{a}\"{a}, T.: Multiresolution
	gray-scale and rotation invariant texture classification with local binary
	patterns. IEEE Transactions on Pattern Analysis and Machine Intelligence
	\textbf{24}(7),  971--987 (2002)
	
	\bibitem{P1901}
	Pearson, F.K.: Liii. on lines and planes of closest fit to systems of points in
	space. The London, Edinburgh, and Dublin Philosophical Magazine and Journal
	of Science  \textbf{2}(11),  559--572 (1901)
	
	\bibitem{PM90}
	Perona, P., Malik, J.: Scale-space and edge detection using anisotropic
	diffusion. IEEE Trans. Pattern Anal. Mach. Intell.  \textbf{12}(7),  629--639
	(1990)
	
	\bibitem{SC20}
	Robert~Singh, K., Chaudhury, S.: {Comparative analysis of texture feature
		extraction techniques for rice grain classification}. {IET Image Processing}
	\textbf{{14}}({11}),  {2532--2540} ({2020})
	
	\bibitem{VZ05}
	Varma, M., Zisserman, A.: A statistical approach to texture classification from
	single images. International Journal of Computer Vision  \textbf{62}(1),
	61--81 (2005)
	
	\bibitem{VZ09}
	Varma, M., Zisserman, A.: A statistical approach to material classification
	using image patch exemplars. IEEE Transactions on Pattern Analysis and
	Machine Intelligence  \textbf{31}(11),  2032--2047 (2009)
	
	\bibitem{VMS20}
	Verma, V., Muttoo, S.K., Singh, V.B.: {Multiclass malware classification via
		first- and second-order texture statistics}. {Computers \& Security}
	\textbf{{97}} ({2020})
	
	\bibitem{VA18}
	Vieira, J., Abreu, E.: Numerical modeling of the two-phase flow in porous media
	with dynamic capillary pressure. Ph.D. thesis, University of Campinas,
	Campinas, SP, Brazil (7 2018)
	
	\bibitem{vieira2020texture}
	Vieira, J., Abreu, E., Florindo, J.B.: Texture image classification based on a
	pseudo-parabolic diffusion model (2020), {A}vailable at
	\url{https://arxiv.org/abs/2011.07173}
	
	\bibitem{W97}
	Weickert, J.: A review of nonlinear diffusion filtering. In: Proceedings of the
	First International Conference on Scale-Space Theory in Computer Vision. pp.
	3--28. SCALE-SPACE '97, Springer-Verlag, Berlin, Heidelberg (1997)
	
	\bibitem{W83}
	Witkin, A.P.: Scale-space filtering. In: Proceedings of the Eighth
	International Joint Conference on Artificial Intelligence - Volume 2. pp.
	1019--1022. IJCAI'83, Morgan Kaufmann Publishers Inc., San Francisco, CA, USA
	(1983)
	
	\bibitem{XJF09}
	Xu, Y., Ji, H., Ferm\"{u}ller, C.: Viewpoint invariant texture description
	using fractal analysis. International Journal of Computer Vision
	\textbf{83}(1),  85--100 (2009)
	
	\bibitem{ZWC20}
	Zhao, G., Wang, X., Cheng, Y.: {Hyperspectral image classification based on
		local binary pattern and broad learning system}. {International Journal of
		Remote Sensing}  \textbf{{41}}({24}),  {9393--9417} ({2020})
	
\end{thebibliography}

\end{document}